\documentclass[a4paper, 11pt, table, xcdraw]{article}
\usepackage[utf8]{inputenc}
\usepackage{graphicx}
\usepackage{amssymb}
\usepackage[preprint]{neurips_2020}

\usepackage{amsmath}

\newcommand{\pluseq}{\mathrel{+}=}
\usepackage{hyperref}

\usepackage{graphicx}
\graphicspath{ {./} }
\usepackage{subcaption}

\usepackage[shortlabels]{enumitem}

\usepackage{wrapfig}

\usepackage{algorithm}
\usepackage{algpseudocode}
\algtext*{EndIf}
\algtext*{EndFor}
\algtext*{EndWhile}

\usepackage[parfill]{parskip}

\usepackage{tikz}
\usetikzlibrary{shapes,arrows}

\usepackage{booktabs}
\usepackage{multirow}
\usepackage{graphicx}
\usepackage{xcolor}
\usepackage{bbm}

\title{Model Evidence with Fast Tree Based Quadrature}

\author{%
  Thomas Foster$^1$\\
  \texttt{thomas.foster@keble.ox.ac.uk} \\
  \And
  Chon Lok Lei$^1$\\
  \texttt{chon.lei@cs.ox.ac.uk} \\
  \And
  Martin Robinson$^1$\\
  \texttt{martin.robinson@cs.ox.ac.uk} \\
  \And
  David Gavaghan$^1$\\
  \texttt{david.gavaghan@dtc.ox.ac.uk} \\
  \And
  Ben Lambert$^2$\\
  \texttt{ben.c.lambert@gmail.com} \\
}

\begin{document}
\maketitle
$^1$Department of Computer Science, University of Oxford; $^2$Department of Global Infectious Disease Analysis, Imperial College London.
\section{Abstract}
High dimensional integration is essential to many areas of science, ranging from particle physics to Bayesian inference. Approximating these integrals is hard, due in part to the difficulty of locating and sampling from regions of the integration domain that make significant contributions to the overall integral. Here, we present a new algorithm called Tree Quadrature (TQ) that separates this sampling problem from the problem of using those samples to produce an approximation of the integral. TQ places no qualifications on how the samples provided to it are obtained, allowing it to use state-of-the-art sampling algorithms that are largely ignored by existing integration algorithms. Given a set of samples, TQ constructs a surrogate model of the integrand in the form of a regression tree, with a structure optimised to maximise integral precision. The tree divides the integration domain into smaller containers, which are individually integrated and aggregated to estimate the overall integral. Any method can be used to integrate each individual container, so existing integration methods, like Bayesian Monte Carlo, can be combined with TQ to boost their performance. On a set of benchmark problems, we show that TQ provides accurate approximations to integrals in up to 15 dimensions; and in dimensions 4 and above, it outperforms simple Monte Carlo and the popular Vegas method \citep{lepage1978new}.

\section{Introduction}
High dimensional integrals appear in diverse areas of science. In particle physics, they are used to calculate scattering amplitudes implicitly represented by Feynman diagrams; in statistical physics and Bayesian inference, they are needed to calculate the partition function or model evidence that normalises the probability distribution over state space. Here, we focus on the types of integral encountered in Bayesian inference, but the approach we develop is equally applicable to more general problems. Specifically, we consider computing weighted integrals of the form,

\begin{equation}\label{eq:partition}
    Z = \int_\Omega f(x)p(x)dx,
\end{equation}

where $\Omega$ is the $D$-dimensional integration domain, $f: \Omega \mapsto \mathbb{R}$, $p$ is a valid probability distribution over $\Omega$ and their product, $f(x)p(x)$, is the integrand. For example, model evidence is computed taking $f(x)$ as the likelihood and $p(x)$ as the prior distribution.

The method presented in this paper, Tree Quadrature (TQ), constructs a regression tree surrogate model of the integrand $f(x)p(x)$. The tree is constructed by recursively partitioning $\Omega$ into a set of smaller containers, which form the leaves of the tree. Containers are then integrated independently, and the partition constructed to optimise accuracy of the overall integral. Tree construction is fast and linear in the dimension of the integration domain, unlike existing surrogate model integration techniques such as Bayesian Quadrature \citep{rasmussen2003bayesian}.

Efficient numerical integration and efficient sampling are highly intertwined -- each method aims to locate and explore areas of high integrand value or high probability mass. Many integration algorithms attempt to solve this problem ``in house''; for example, the Vegas and importance sampling algorithms discussed in \S\ref{sec:existing_methods} do this. Rather than reinvent the wheel, TQ outsources sampling to whatever method is favourable for a particular integrand, so it can use any existing class of sampling algorithm, including Markov chain Monte Carlo (MCMC) (e.g. for high dimensional posteriors) and nested sampling \citep{skilling2006nested} (e.g. for multimodal posteriors). Sampling need not even be done by samplers; any method that allows exploration of the integrand works with TQ, allowing users the freedom to incorporate prior knowledge of where it is best to sample next. As such, rounds of ``active sampling'' -- where the integrand is explored in areas of high inaccuracy -- can be incorporated to further refine tree structure. 

We now present a brief overview of existing integration techniques to provide motivation for TQ. Note that the field of numerical integration is large, and here we only summarise those approaches most relevant to TQ.

\subsection{Existing integration methods}\label{sec:existing_methods}
The computation used by many approximate integration techniques can be expressed as a quadrature rule,

\begin{equation}
Z \approx \sum_{i=1}^{N} w_i f(x_i)p(x_i),
\end{equation}
where $x_1, \ldots, x_N$ are function evaluation points and $w_1, \ldots, w_N$ are corresponding weights. Techniques differ in (1) how they select the locations of these function evaluation points, and (2) how they choose weights for each point.

Early quadrature rules selected function evaluation points deterministically, often by gridding the integration domain. For example, the trapezium rule evaluates the integrand at points $x_1, \ldots, x_N$ spaced a distance $\Delta$ apart on a uniform grid. It then combines values of the integrand at these points using precomputed weights of $w_1 = w_N = \Delta / 2$ and $w_i = \Delta $ otherwise. Such techniques are, however, infeasible in more than a handful of dimensions because the number of function evaluation points needed to uniformly ``grid'' $\Omega$ increases exponentially with integral dimensions.

To tackle higher dimensions, modern methods, instead, draw evaluation points randomly from some distribution. The Simple Monte Carlo integrator (SMC) \citep{lambert2018Problems}, for example, computes,

\begin{equation}\label{eq:smc}
    Z \approx \frac{1}{N}\sum_{i=1}^{N} f(x_i),
\end{equation}

where $x_i \stackrel{\text{i.i.d.}}{\sim} p(x)$. This is a quadrature rule with $w_i = \frac{1}{N p(x_i)}$. 

In practice, the distribution used to draw function evaluation points should be tailored to the integrand because often only a small subspace of the integration domain contributes to the integral. Function evaluation points selected from generic distributions can completely miss such subspaces, resulting in underestimates. ``Importance sampling'' methods (see, for example, \cite{lambert2018Introduction}) instead aim to generate then sample from bespoke probability distributions that allow for more adequate exploration of the integrand. In importance sampling, eq. \eqref{eq:partition} is rewritten as,

\begin{equation}\label{eq:importance_sampling}
    Z = \int_\Omega f(x)\frac{p(x)}{g(x)}g(x)dx,
\end{equation}

where $g(x)$ is the ``importance distribution'' that can be independently sampled from. Eq. \eqref{eq:importance_sampling} leads to the alternative estimator,

\begin{equation}
    Z \approx \frac{1}{N}\sum_{i=1}^{N} f(x_i)\frac{p(x_i)}{g(x_i)},
\end{equation}

where $x_i \stackrel{\text{i.i.d.}}{\sim} g(x)$. Choice of $g(x)$ then yields different estimators of $Z$, and this choice can dramatically affect estimator variance; $g(x) = f(x)p(x)/Z$ is optimal, but in general it is not possible to independently sample from this distribution. 

A common approach is to construct, and then sample from, a series of importance distributions that iteratively approach optimality. See, for example, the thermodynamic integration approach in \cite{neal2001annealed}), or the Vegas method \citep{lepage1978new}, popular in particle physics (amongst other similar routines implemented in the ``Cuba'' numerical integration package \citep{hahn2005cuba}). In Vegas, the integration domain is decomposed into discrete hyper-cuboidal bins, whose dimensions are refined in rounds of importance sampling to maximise approximation accuracy. One crucial assumption of Vegas is that the underlying integrand has a structure permitting it to be factorised into a product of densities along each coordinate axis. This factorisation allows the number of bins needed to grow linearly, rather than exponentially, with the problem dimension.

A flaw in many modern stochastic integration techniques, including the Vegas and Importance Sampling algorithms discussed above, is that evaluation points are selected independently and do not account for where the integrand has already been evaluated \citep{o1987monte}. This leads to inefficient exploration of the integration domain, with the integrand being over-sampled in small, localised regions and under-sampled elsewhere. More advanced techniques, such as the surrogate model techniques discussed next, can adjust the distribution used to sample function evaluation points on the fly to achieve more efficient coverage of the integration domain.

Surrogate model based techniques aim to approximate the integrand itself by fitting a simple-to-integrate function to the evaluation points. By including prior knowledge about how the integrand varies in the calculation, surrogate models effectively reduce the variance of integral estimates. They can also be more efficient with their function evaluations than non-model-based integration. By examining the fitted model, they can identify areas of uncertainty or disproportionate contribution to the integral. This permits active sampling of evaluation points allowing more efficient exploration of the integration domain. 

Bayesian Monte Carlo (BMC) \citep{rasmussen2003bayesian} is a popular numerical integration method based on fitting a Gaussian Process (GP) to function evaluations to build a surrogate model. Specifically, under a GP prior, the joint distribution of finite function evaluations is assumed Gaussian,

\begin{equation}
    \boldsymbol{f}_{1:n} := (f(x_1), f(x_2),...,f(x_n))^T \sim \mathcal{N}(0, K),
\end{equation}

where $K$ is a covariance matrix dictated by a given choice of covariance function, ${K_{ij}=\text{Cov}(f(x_i), f(x_j)) =k(x_i, x_j),\;\forall i, j}$. BMC is a Bayesian approach that uses function evaluations $\mathcal{D}:=(f(x_1), f(x_2),...,f(x_n))$ to derive an (infinite dimensional) posterior surrogate model of the function, $\boldsymbol{f}|\mathcal{D}$. BMC treats the value of the integral, $Z$, in eq. \eqref{eq:partition} as an uncertain quantity and seeks to update the probability distribution describing this uncertainty, $p(Z|\mathcal{D})$; as such, an estimate of $Z$ is obtained via its posterior mean, $\mathrm{E}_{f|\mathcal{D}}(Z)$. Any sampling method, including MCMC and active sampling, can be used to generate $\mathcal{D}$, although the computational cost of doing inference for GPs scales as $n^3$ due to costs associated with constructing and manipulating the covariance matrix.

\begin{figure}[h]
\includegraphics[width=\linewidth]{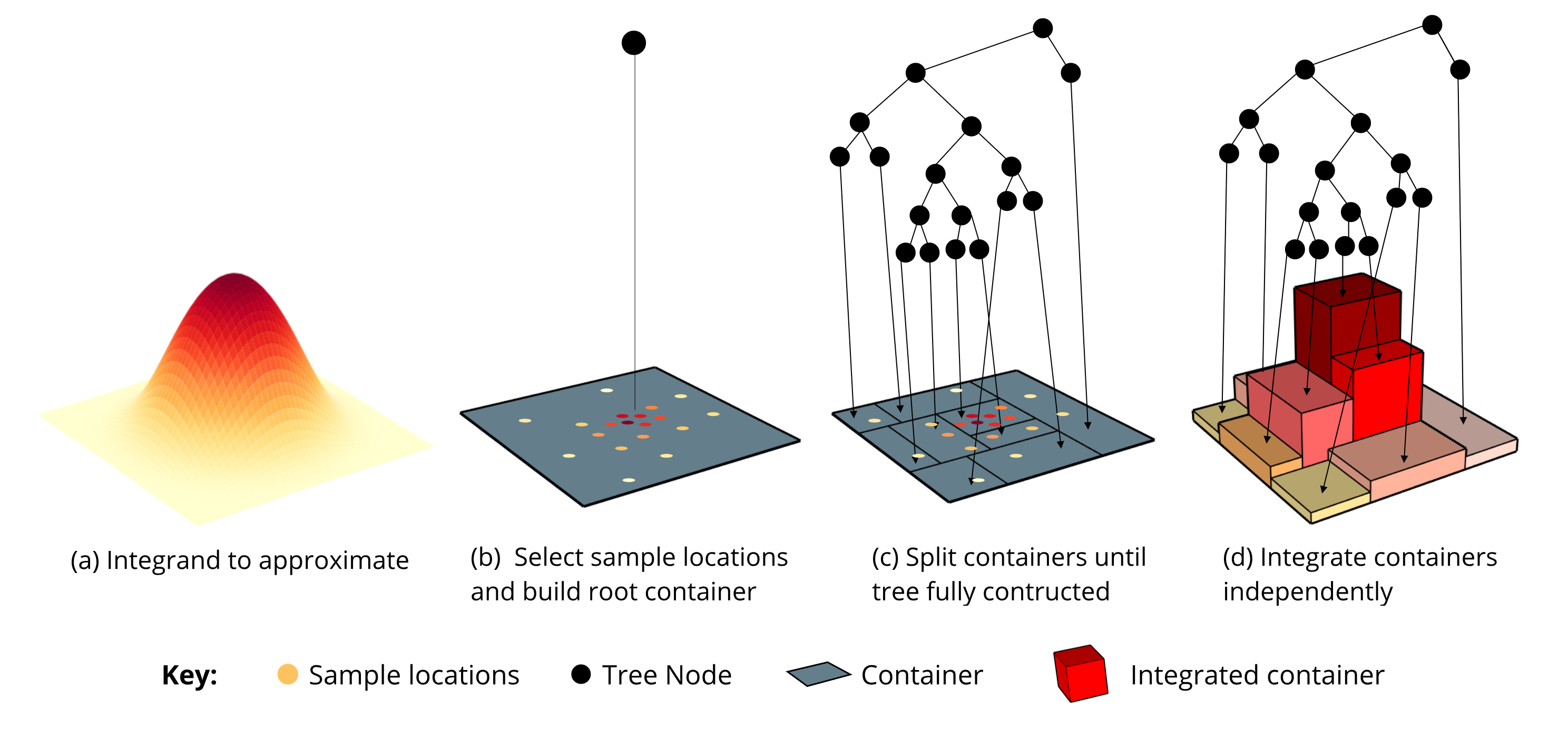}
\caption{\textbf{Construction of the regression tree.} (a) shows the integrand; as the algorithm progresses (from panels (b)-(d)), the integration domain is repeatedly partitioned into sub-domains, each with its own contribution to the overall integral. The partitioning may terminate when the contributions of the sub-domains are largely similar to one another.}
\label{tree construction}
\end{figure}

\section{Tree Quadrature}

The surrogate model used by TQ is the regression tree, a hierarchical structure of linked nodes, visualised in Fig. \ref{tree construction}. Each node in the tree is a data structure, called a ``container'', that represents a finite subset of the original integration domain. A container stores any  samples of the function that fall within it. We differentiate between the location of the samples, and the value of the function at these locations. For a container $c$, we denote its sample locations $c.X$ and function values $c.Y$.

There are two fundamental operations that can be performed on a container, $c$: it can be split and it can be integrated. Splitting a container partitions the region of space it represents, and produces two smaller child containers to represent this partition. The details of container splitting, and how a partition is computed to maximise the overall integral accuracy, are discussed in section \ref{optimal splitting}. Integrating a container involves using $c.X$ and $c.Y$ to produce an estimate for the integral over the region of space that $c$ represents.

TQ is general framework that allows much flexibility in how these operations are employed to produce a final integral. We now discuss two implementations of this framework, simple TQ (TQ-s) and active TQ (TQ-a).

\begin{algorithm}[t!]
\begin{algorithmic}
\State 
\State \textbf{Input:} integrand $f: \mathbb{R^D}\mapsto\mathbb{R}$
\State \qquad \quad integration domain $\Omega \subset\mathbb{R^D}$
\State \qquad \quad sample locations $X_0 \in \mathbb{R}^{N_0\times D}$
\State \qquad \quad function values $Y_0  \in {R}^{N_0}$
\State
\State root $\gets$ Container$($domain: $\Omega$, $X$: $X_0$, $Y$: $y_0)$
\Comment{construct tree}
\State $Q \gets$ Queue$($root$)$
\State
\While{exists cont $\in Q$ with stopping\_condition(cont) = False}
    \Comment{%
        \smash{
            \parbox[t]{.25\linewidth}{a high accuracy choice is to stop when nodes contain only 1 sample}
        }
    }
    \State 
    \State cont $\gets$ $Q$.pop$()$
    \State $\text{child}_1$, $\text{child}_2$ $\gets$ split$(\text{cont})$
     \Comment find optimal split of node
    \State $Q\gets Q\,\,\cup \{\text{child}_1, \text{child}_2  \}$
\EndWhile
\State
\State $I$ = 0
\Comment{integrate tree}
\For{cont $\in Q$}
    \State $I$ $\pluseq$ integrate$($cont$)$ \Comment integrate container via chosen method
\EndFor
\State \textbf{Return} $I$
\end{algorithmic}
\caption{Simple Tree Quadrature}
\label{stq}
\end{algorithm}

\subsection{Simple TQ (TQ-s)}
\label{tq-s}

TQ-s is the most basic form of TQ and is outlined in Algorithm \ref{stq}. Its inputs are a pre-generated set of sample locations, $X_0$, and function values $Y_0$, which TQ-s then fits a regression tree to. TQ places no requirements on the statistical distribution that members of $X_0$ are drawn from; any method may be used, provided it results in adequate coverage of the integrand.

$X_0$ and $Y_0$ are then placed inside a root container representing the entire integration domain. TQ-s maintains a queue, $Q$, of containers and splits them until they all satisfy a given stopping condition. The exact stopping condition used determines the overall algorithm run-time since it determines the number of splits that are performed and therefore the depth of the tree constructed. Whilst containers can be split until they contain just 1 sample, we found in practise it is not always necessary to construct such a deep tree. For example, splitting containers in areas of low function curvature (such as in the tails of the integrand in Fig.\ref{tree construction}) offers minimal improvements in accuracy: it's typically more efficient to stop when the variance of a container is below a specified threshold.

Once all containers satisfy the stopping condition, they are integrated. The overall integral is the sum of these individual integrals. The goal of TQ is to partition the integration domain into smaller containers, over which the integrand behaves more manageably. This means any existing integration method is likely to be more effective when applied to individual containers.

A basic container integration method is to multiple a container's volume with a representative value of the integrand over that domain. This approach is computationally inexpensive, but it can be very sensitive to how the representative value is calculated -- using the mean or median of $c.Y$ proved to be unstable in higher dimensions, as did evaluating the function at the midpoint of the container. We found that taking the mean of a small number ($\sim 10$) of further function evaluations, uniformly sampled over the area of the container, was a more stable approach, and it is used for the results presented at the end of this paper. We refer to this approach as the ``random container integral''.

In the final regression tree, each container contains relatively few samples, which reduces the computational cost of using an integration method over that container. For example, BMC has a runtime of $O(n^3)$ when fitting a Gaussian process (GP) to $n$ function evaluations. Fitting a GP to containers individually can be much faster than fitting a single GP to all function evaluation points over the entire integration domain. However, integrating the fitted GP over the finite region the container represents is not trivial, and we leave this to future work.

\subsection{Active TQ (TQ-a)}
TQ-a extends TQ-s with active sampling techniques. It begins in the same way as TQ-s, by constructing a root container to hold the provided $X_0$ and $Y_0$, and splitting containers until they all satisfy a given stopping condition. The difference is that, after this, instead of immediately integrating the resulting containers, TQ-a has an extra phase of tree construction to further refine the model. The containers from the previous phase are loaded into a priority queue, where they are ordered according to the estimated inaccuracy of their integrals. 

The most challenging part of TQ-a is estimating the uncertainty in a container's integral. We found that using the range of $c.Y$ was a good heuristic to do this. In future work, we wish to explore Bayesian approaches to do this.

TQ-a repeatedly selects the container with the greatest inaccuracy from the queue, and selects one new sample location in the region of space that the container represents. We drew this location uniformly at random over the container, but future extensions of TQ-a could use information about $c.X$ and $c.Y$ to do this more intelligently. The function is then evaluated at this location, and the container is split using this new information. Its children are added to the priority queue, and the process repeats until TQ-a has used up its budget of further function evaluations.

There are numerous benefits of TQ-a. It offers higher integral accuracy, since areas of model uncertainty are systematically eliminated, and it allows TQ to work even when a limited number of initial samples are provided. By choosing function evaluations points in areas where they are most needed, it is also more judicious -- especially useful when function evaluation is costly.

\begin{figure}[h!]

    \centering
    \begin{subfigure}[t]{\linewidth}
        \begin{subfigure}[t]{0.3\linewidth}
            \centering
            \includegraphics[width=\linewidth]{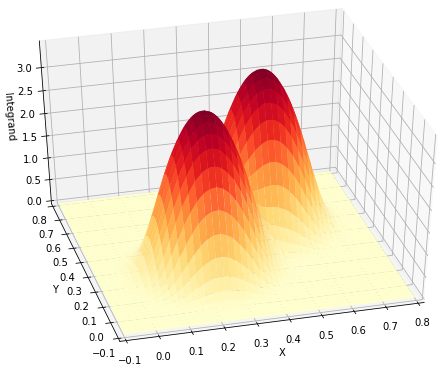}
            \caption{Camel Problem Integrand surface}
            \vspace{0.45cm}
        \end{subfigure}
        \hfill
        \begin{subfigure}[t]{0.3\linewidth}
            \centering
            \includegraphics[width=\linewidth]{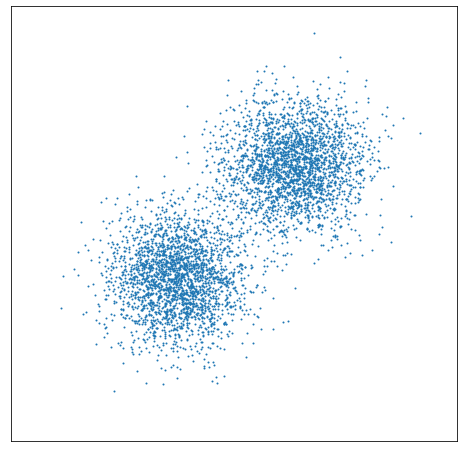}
            \caption{The preselected samples used for the following runs of TQ}
            \vspace{0.45cm}
        \end{subfigure}
        \hfill
        \begin{subfigure}[t]{0.3\linewidth}
            \centering
            \includegraphics[width=\linewidth]{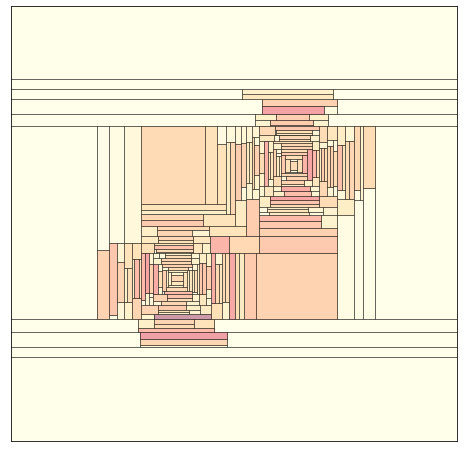}
            \caption{Min SSE Axial Split}
            \vspace{0.85cm}
        \end{subfigure}
    \end{subfigure}
    \hfill
    \begin{subfigure}[t]{\linewidth}
        \hfill
        \begin{subfigure}[t]{0.3\linewidth}
            \centering
            \includegraphics[width=\linewidth]{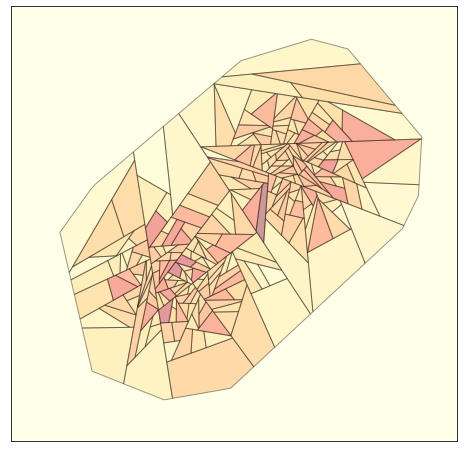}
            \caption{Random Split}
        \end{subfigure}
        \hfill
        \begin{subfigure}[t]{0.3\linewidth}
            \centering
            \includegraphics[width=\linewidth]{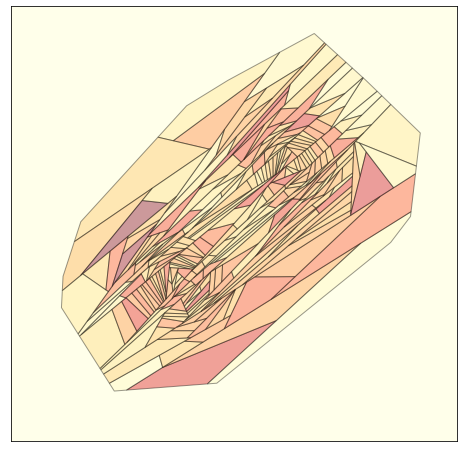}
            \caption{Min SSE Split}
        \end{subfigure}
        \hfill
        \begin{subfigure}[t]{0.3\linewidth}
            \centering
            \includegraphics[width=\linewidth]{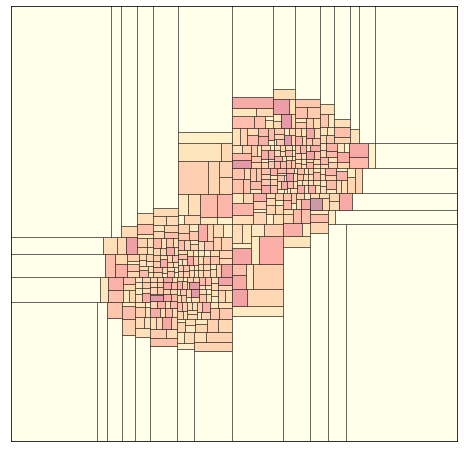}
            \caption{KD Split}
        \end{subfigure}
    \end{subfigure}
    \caption{\textbf{Camel distribution and example regression trees fitted to independent samples.} (a) shows the Camel distribution PDF defined in \S\ref{sec:results}; (b) shows independent samples from this distribution; and (c)-(f) show regression tree models fitted to these samples, each using different splitting methods.}
    \label{fig:splits}
\end{figure}

\subsection{Splitting containers}
\label{optimal splitting}

In Fig. \ref{fig:splits}, we illustrate how changing the method used to split containers yields different decompositions of the domain. In Figs. \ref{fig:splits}(a)\&(b), we show an example integrand and independent samples from it to which different splitting methods are trialled. In this paper, we split containers using only half-planes, resulting in convex sub-domains as seen in Fig. \ref{fig:splits}(c)-(f). If TQ is restricted further to half-planes that run parallel to a coordinate axis, so-called ``axial'' splits, then containers represent hyper-rectangles of space, as seen in Fig. \ref{fig:splits}(c). These restrictions make it easier to compute the split for each container that will maximise the accuracy of the overall integral. We define an optimal split as one which minimises the sum of a metric over the child containers it produces. A sensible choice for this metric is the variance of $c.Y$, which we denote as the ``MinSSE'' splitting rule.

Finding a half-plane that minimises this metric is straightforward for axial-splits. For a container with $N$ function evaluation points in $D$ dimensions, the number of candidate half-planes is $N \times D$, and thus the optimal split can be found by exhaustive search. The rule that performs this optimisation is known as the MinSSE axial splitting rule and is visualised in Fig. \ref{fig:splits}(c).

Without the axial restriction, the set of candidate half-planes is much larger, making exhaustive search intractable. We found that creating a test pool of randomly drawn half-planes, and selecting the best of these approximated optimal splitting adequately. The quality of this approximation depends on the size of the test pool: Fig. \ref{fig:splits}(d) was produced with a test pool of size 1, whereas Fig. \ref{fig:splits}(e) used 200 such tests and nicely follows the contours of the integrand. Whilst non-axial splitting typically produces containers more aligned with function contours, it is harder to manage and integrate convex-polytopes in more than 5-6 dimensions, and we leave their analysis to future work.

We also considered using the KD splitting rule, which is less computationally intensive than MinSSE. It first finds the axis, d $\in \{1, \ldots, D\}$, with the highest variance of $c.X$. It then splits perpendicular to this axis (see \ref{fig:splits}(f) for an example KD-decomposition). There are just $D$ candidate KD splits, making this significantly faster than MinSSE in practise.

\section{Results}\label{sec:results}

\begin{figure}
    \centering
    \begin{subfigure}[t]{0.49\linewidth}
            \centering
            \includegraphics[width=\linewidth]{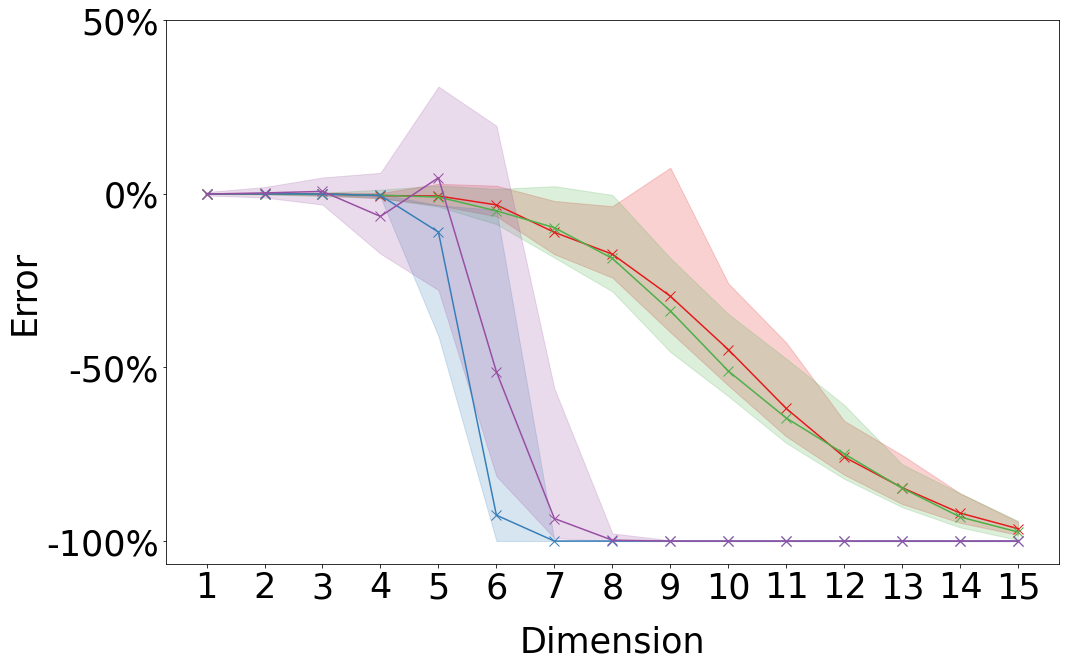}
        \end{subfigure}
        \hfill
        \begin{subfigure}[t]{0.49\linewidth}
            \centering
            \includegraphics[width=\linewidth]{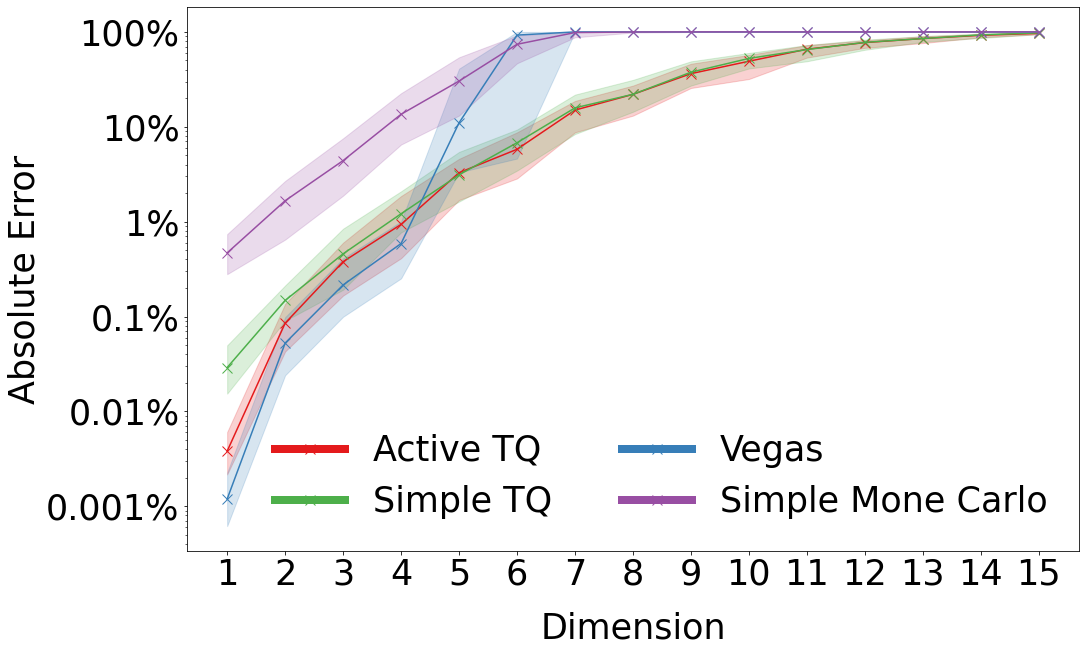}
    \end{subfigure}
    \caption{\textbf{Integrating the Camel distribution in varying dimensions.} Solid lines are medians over 100 replicates; shaded regions show the interquartile ranges. \textsuperscript{\ref{footnote:hyperparameters}}}
    \label{fig:results}
\end{figure}

\begin{table}[h!]
\caption[caption for lof]{\textbf{Method performance for benchmark problems.}\protect\footnotemark}
\label{table:results}
\resizebox{\textwidth}{!}{%
\texttt{
\begin{tabular}{@{}llrrr@{}}
\toprule
{\color[HTML]{000000} }                                   & {\color[HTML]{000000} }                                  & \multicolumn{3}{c}{{\color[HTML]{000000} \textbf{Median percentage error +- standard deviation}}}                                                                                                      \\
\multirow{-2}{*}{{\color[HTML]{000000} \textbf{Problem}}} & \multirow{-2}{*}{{\color[HTML]{000000} \textbf{Method}}} & \multicolumn{1}{c}{{\color[HTML]{000000} \textbf{1 Dimension}}} & \multicolumn{1}{c}{{\color[HTML]{000000} \textbf{5 dimensions}}} & \multicolumn{1}{c}{{\color[HTML]{000000} \textbf{10 dimensions}}} \\ \cmidrule(r){1-5}
{\color[HTML]{000000} \textbf{Gaussian}}                  & {\color[HTML]{000000} Simple Monte Carlo}                & {\color[HTML]{000000} -0.03228 +-   1.16659}                    & {\color[HTML]{000000} -12.54548 +-  \hspace{1ex}70.54963}                    & {\color[HTML]{000000} -100.00000 +-  \hspace{1ex}20.10814}                    \\
{\color[HTML]{000000} }                                   & {\color[HTML]{000000} Vegas}                             & {\color[HTML]{000000} -0.00001 +-   0.00115}                    & {\color[HTML]{000000} -11.01276 +-  \hspace{1ex}42.61782}                    & {\color[HTML]{000000} -100.00000 +-   \hspace{1ex}\hspace{1ex}0.00000}                    \\
{\color[HTML]{000000} }                                   & {\color[HTML]{000000} TQ-s}                         & {\color[HTML]{000000} -0.01359 +-   0.05232}                    & {\color[HTML]{000000} -1.01358 +-   \hspace{1ex}\hspace{1ex}6.37404}                     & {\color[HTML]{000000} -59.78854 +- 176.30427}                     \\ 
{\color[HTML]{000000} }                                   & {\color[HTML]{000000} TQ-a}                         & {\color[HTML]{000000} -0.00034 +-   0.00571}                    & {\color[HTML]{000000} -0.38929 +-   \hspace{1ex}\hspace{1ex}6.12029}                     & {\color[HTML]{000000} -56.61069 +-  \hspace{1ex}81.12971}                     \\ \cmidrule(r){1-5}
{\color[HTML]{000000} \textbf{Camel}}                     & {\color[HTML]{000000} Simple Monte Carlo}                & {\color[HTML]{000000} -0.06432 +-   0.64505}                    & {\color[HTML]{000000} 4.64382 +-  \hspace{1ex}53.53355}                      & {\color[HTML]{000000} -100.00000 +-   \hspace{1ex}\hspace{1ex}0.64222}                    \\
{\color[HTML]{000000} }                                   & {\color[HTML]{000000} Vegas}                             & {\color[HTML]{000000} -0.00051 +-   0.00180}                    & {\color[HTML]{000000} -11.01175 +-  \hspace{1ex}35.03981}                    & {\color[HTML]{000000} -100.00000 +-   \hspace{1ex}\hspace{1ex}0.00000}                    \\
{\color[HTML]{000000} }                                   & {\color[HTML]{000000} TQ-s}                         & {\color[HTML]{000000} -0.01395 +-   0.05855}                    & {\color[HTML]{000000} -0.92340 +-   \hspace{1ex}\hspace{1ex}7.32393}                     & {\color[HTML]{000000} -51.01571 +-  \hspace{1ex}59.93169}                     \\
{\color[HTML]{000000} }                                   & {\color[HTML]{000000} TQ-a}                         & {\color[HTML]{000000} 0.00048 +-   0.00611}                     & {\color[HTML]{000000} -0.62759 +-   \hspace{1ex}\hspace{1ex}8.33251}                     & {\color[HTML]{000000} -44.84099 +-  \hspace{1ex}47.27089}                     \\ \cmidrule(r){1-5}
{\color[HTML]{000000} \textbf{Quad}}                      & {\color[HTML]{000000} Simple Monte Carlo}                & {\color[HTML]{000000} -0.15660 +-   1.25727}                    & {\color[HTML]{000000} -100.00000 +- 160.14255}                   & {\color[HTML]{000000} -100.00000 +-   \hspace{1ex}\hspace{1ex}0.00000}                    \\
{\color[HTML]{000000} }                                   & {\color[HTML]{000000} Vegas}                             & {\color[HTML]{000000} -100.00000 +-   0.00000}                  & {\color[HTML]{000000} -99.94417 +-   \hspace{1ex}\hspace{1ex}7.69237}                    & {\color[HTML]{000000} -100.00000 +-   \hspace{1ex}\hspace{1ex}0.00000}                    \\
{\color[HTML]{000000} }                                   & {\color[HTML]{000000} TQ-s}                         & {\color[HTML]{000000} -0.03408 +-   0.14225}                    & {\color[HTML]{000000} -12.67292 +-  \hspace{1ex}23.44511}                    & {\color[HTML]{000000} -94.61277 +-  \hspace{1ex}17.83062}                     \\
{\color[HTML]{000000} }                                   & {\color[HTML]{000000} TQ-a}                         & {\color[HTML]{000000} 0.00004 +-   0.01890}                     & {\color[HTML]{000000} -8.17305 +-  \hspace{1ex}15.56113}                     & {\color[HTML]{000000} -90.66218 +- 708.67058}                     \\ \bottomrule
\end{tabular}%
}}
\end{table}
\footnotetext{All methods use 12k total samples in total, with TQ-a requiring 25\% less of these to be provided to it initially. Vegas and Simple Monte Carlo use their own sampling techniques to draw these. TQ methods used samples drawn from the posterior distribution. All TQ methods use the Min SSE Axial split method and the random container integral approach (see section \ref{tq-s}). Quantiles calculated from 100 replicates.\label{footnote:hyperparameters}}

We tested TQ on three example problems of the same form as eq. \eqref{eq:partition}. Fig. \ref{fig:results} compares the performance of four algorithms, Simple Monte Carlo, Vegas, TQ-s and TQ-a, on one of these problems, the Camel distribution. This takes $f(x) = \mathcal{N}(x|\mu_1, \Sigma) + \mathcal{N}(x|\mu_2, \Sigma)$, the sum of two multivariate normal distributions, with $\mu_1, \mu_2$ placed at $\frac{1}{3}$ and $\frac{2}{3}$ along the unit diagonal; $\Sigma=\frac{1}{200}\textbf{I}$ is a diagonal matrix. The integration domain is the unit hyper-cube $\{0, 1\}^D$, and the prior, $p$, is the uniform distribution over this range. All four algorithms are allowed 12,000 function evaluations each. Fig. \ref{fig:results}(a) shows the error of the algorithms on a linear scale, and Fig. \ref{fig:results}(b) shows the absolute error of the algorithms on a log scale.

The performance of all algorithms worsens as problem dimension increases: the linear error, shown in (a), increases from ca.0\% error to a ca.100\% underestimate. SMC's performance worsens rapidly with increasing dimension. Vegas has the smallest absolute error in lower dimensions, but, like SMC, struggles in even moderately high dimensions. Whilst providing slightly coarser estimates in lower dimensions, TQ-s and TQ-a generalise much better to higher ones.

Table \ref{table:results} summarises the performance of the same algorithms over the Camel problem and two further problems: the Gaussian distribution and the Quad distribution. The Gaussian distribution is a multivariate normal distribution centered at the origin (specifically, $f(x) = \mathcal{N}(x|0,\, \frac{1}{200}\textbf{I})$). The integration domain is the hyper-cube $\{-1, 1\}^D$, and $p$ is the uniform distribution over this. The Quad Camel PDF is the sum of 4 multivariate normal distributions placed at 2, 4, 6, and 8 units along the diagonal of the hyper-cube $\{0, 10\}^D$; the Quad distribution has the same covariance matrix as in Gaussian and Camel, $\frac{1}{200}\textbf{I}$, but the larger integration domain, $\{0, 10\}^D$, means the concentration of probability mass is tiny compared with the integral limits. Table \ref{table:results} shows the median percentage error of 100 runs of each algorithm. In all bar the Gaussian problem in one dimension, TQ-a has the smallest median percentage error, followed closely by TQ-s. For the Gaussian and Camel problems in one dimension, Vegas is competitive with TQ. For the Quad problem, Vegas consistently fails to find areas of high probability mass and underestimates.

\section{Diagnosing unreliable integrations}

\begin{figure}[htb!]
    \includegraphics[width=\linewidth]{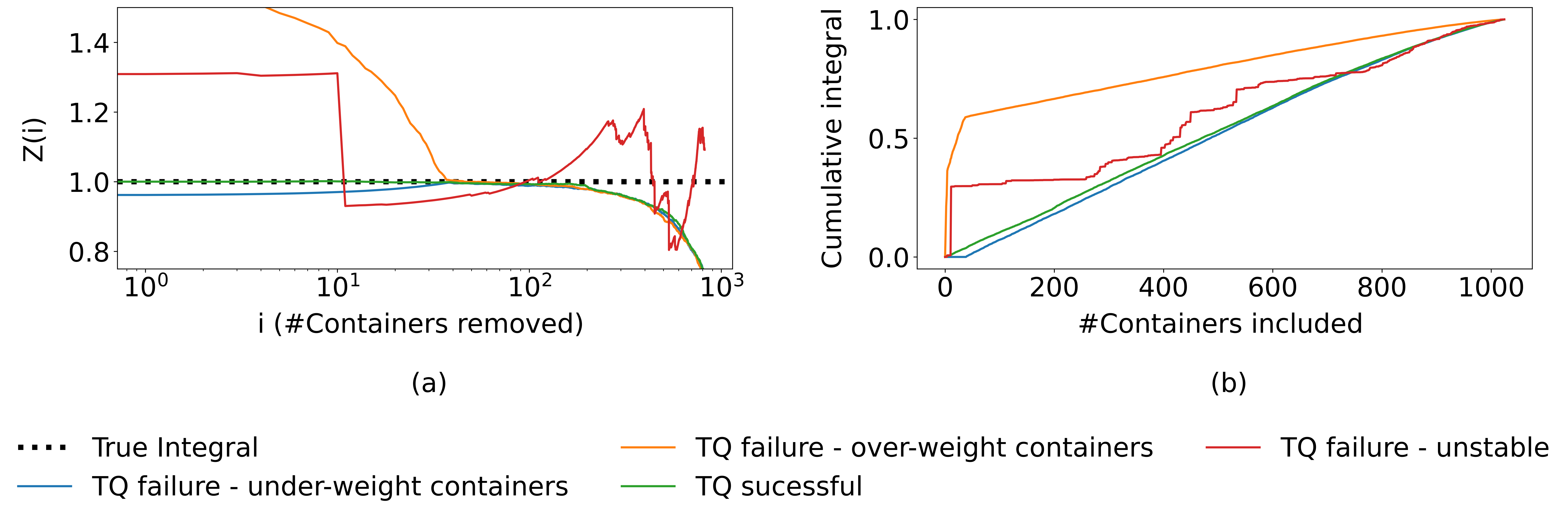}
    
    \caption{\textbf{Diagnosing problems with TQ integration.} The four solid lines are runs of TQ on various problems introduced in \S\ref{sec:results}. (a) uses eq. \eqref{eq:reweighting_diagnostic} to re-estimate the integral as containers are removed from the model, largest first, and the dashed line shows the true integral value; (b) plots the cumulative integral value as more containers are included, again starting with the largest.}
    \label{fig:diagnostic}
\end{figure}

TQ, like all other approximate integration methods, can fail, with the failure rate increasing along with integral dimensions. It is important to be able to recognise such failures; partly to avoid embarrassment but also because, often, adjustments can be made that alleviate the problem. In this section, we outline a series of diagnostic approaches that can be run after integration with TQ to identify instabilities in the result.

The first of these diagnostic methods is specific to calculation of model evidence in Bayesian inference (or, equally, the partition function in statistical mechanics). It requires that input samples used to construct the regression tree are generated by sampling from the posterior distribution. Bayes' rule for inference can be rearranged then integrated as,

\begin{equation}\label{eq:subset_integral}
    Z = \frac{\int_{\omega\subseteq\Omega} p(\mathcal{D}|x) p(x) dx}{\int_{\omega\subseteq\Omega} p(x|\mathcal{D})dx},
\end{equation}

where $x$ is a parameter vector; $p(x|\mathcal{D})$ is the posterior; $p(\mathcal{D}|x)$ is the likelihood; $p(x)$ is the prior; and $\omega$ is either the whole integration domain, $\Omega$, or a subset of it. Because eq. \eqref{eq:subset_integral} holds irrespective of $\omega$, $Z$ can be calculated by using integrals over any subset of the full domain. If $\omega=\Omega$, in other words, the integral extends over the entire parameter domain, the integral of the posterior density -- the denominator of eq. \eqref{eq:subset_integral} -- is 1, and the standard rule integral for calculating the evidence is recapitulated. If, instead, $\omega\subset\Omega$, the posterior integral $\int_{\omega} p(x|\mathcal{D})dx < 1$, and $\int_{\omega\subseteq\Omega} p(\mathcal{D}|x) p(x) dx<Z$, but their ratio still yields correct calculation of $Z$.

\begin{wrapfigure}{L}{0.3\textwidth}
\centering
\includegraphics[width=0.3\textwidth]{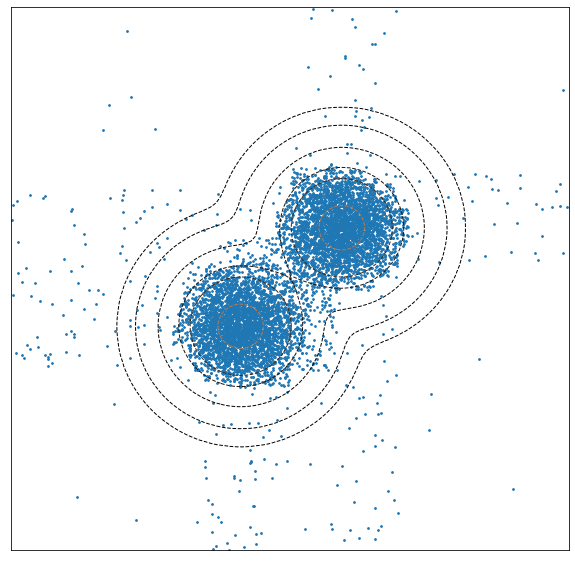}
\captionsetup{width=0.3\textwidth}
\caption{\textbf{Posterior samples from fitted regression tree for the Camel problem.} Here, 8k samples were drawn; the model was fitted using minSSE axial and containers integrated using the median of $c.Y$. \label{fig:frog1}}
\end{wrapfigure}

Eq. \eqref{eq:subset_integral} motivates a family of methods for approximating the evidence, with each member corresponding to a different integral domain, $\omega$. Specifically, TQ can approximate the numerator of eq. \eqref{eq:subset_integral} across any such domain. An unbiased estimator of the denominator is then given by the proportion of posterior samples included in $\omega$ since $Z_1 = \int_{\Omega} \mathbbm{1}(x\in\omega) p(x|\mathcal{D})dx$, where $\mathbbm{1}(a)$ is the indicator function equal to 1 if $a$ is true; 0 otherwise. Hence, the denominator of eq. \eqref{eq:subset_integral} can be estimated using,

\begin{equation}\label{eq:reweighting_diagnostic}
    Z_1 = \mathbb{E}(\mathbbm{1}(x\in\omega)) \approx \frac{1}{N}\sum_{i=1}^{N} \mathbbm{1}(x_i\in\omega),
\end{equation}

where $x_i\sim p(x|\mathcal{D})$. This means that, in theory, we are free to choose whichever subdomain $\omega$ desired and can use eq. \eqref{eq:reweighting_diagnostic} to obtain an unbiased estimator of $Z_1$, leading to an estimate of $Z$ through eq. \eqref{eq:subset_integral}. In practice, the smaller we choose the subdomain to be, the larger the variance of estimates of $Z_1$ and $Z$.

We use this freedom to choose $\omega$ to determine the robustness of $Z$ calculation via TQ. To do this, we first estimate $Z$ using the full integration domain, $\Omega$; in doing so, we fit the regression tree surrogate model to posterior samples, resulting in a partitioning of the domain into a collection of containers. We then iteratively remove the largest containers and note the proportion of posterior samples remaining in doing so. Taking the ratio of the sum of integrals of the remaining containers to this proportion, this yields a new estimate of $Z(i)$, where $i$ is the number of containers removed. By plotting the number of containers removed versus $Z(i)$, we visualise a series of integral estimates: in the left panel of Fig. \ref{fig:diagnostic}, we plot this diagnostic for four TQ runs on various problems defined in \S\ref{sec:results}.

If TQ is working as desired, $Z(i)$ is relatively stable (green line in left panel of Fig. \ref{fig:diagnostic}). If the results are highly sensitive to the number of containers removed, this indicates that particular boxes are influential for the calculation, and the calculation result is unreliable. The left panel of Fig. \ref{fig:diagnostic} shows three examples of this. If the largest, outer, containers have underestimated integrals, $Z(i)$ increases towards the true value as they are removed, as shown by the blue line (the converse is true for underestimates -- see orange line). If various regions have miscalculated integrals, $Z(i)$ is unstable, as for the red line.

A similar approach for determining the fragility of TQ's integral estimate is to assess contributions of individual containers. In the right panel of Fig. \ref{fig:diagnostic}, we plot the number of containers included versus the cumulative integral value over these containers for the same TQ runs as for the left panel. We include the largest containers first. In this plot, sharp jumps in the integral value indicates sensitivity to regions of parameter space. A benefit of this diagnostic approach is that it can be applied to TQ methods that use function evaluation points drawn from any distribution.

A regression tree surrogate model of the integrand can also be used to generate (approximate) samples from it. By randomly selecting a container with probability proportional to its contribution to the integral, this effectively generates samples in proportion to the probability mass contained within it. To generate samples within a container, any sampling method can be used, although, if volumes are all comparably small, uniform sampling within them provides a reasonable approximation. Plotting these can then often highlight, for example, discontinuities in the surrogate regression tree that hinder accuracy. Fig. \ref{fig:frog1} shows this for a model fitted to the Camel distribution that over-weights the values of outer containers. 

\section{Conclusion}
TQ provides a framework to separate sampling from the process of using those samples for integration. It provides efficient and scalable ways to divide the integration domain into smaller sub-domains, each of which can be integrated independently using any existing integration approach. Our results suggest that TQ is more accurate in higher dimensions than existing methods. The implementation is available at https://github.com/thomfoster/treeQuadrature.

\section*{Contributions}
MR conceived the initial idea for this paper based on KD-trees, which TF took and produced an initial proof of concept. TF then conceived of using regression trees instead (which proved more effective), and, with BL, CLL, MR and DG, developed most theory in this paper. TF designed the code structure for TQ and wrote the code. BL conceived the diagnostic approach for evaluating integration performance. TF and BL drafted the original paper. TF, BL, CLL, MR and DG read and edited the paper and contributed throughout its development. 

\bibliography{integral_references}

\begin{thebibliography}{}

\bibitem[\protect\citename{Hahn, }2005]{hahn2005cuba}
Hahn, T. 2005.
\newblock Cuba—a library for multidimensional numerical integration.
\newblock {\em Computer Physics Communications}, {\bf 168}(2), 78--95.

\bibitem[\protect\citename{Lambert, }2018a]{lambert2018Introduction}
Lambert, B. 2018a.
\newblock An introduction to importance sampling.
\newblock {\em YouTube}.

\bibitem[\protect\citename{Lambert, }2018b]{lambert2018Problems}
Lambert, B. 2018b.
\newblock The problems with using simple Monte Carlo to determine the marginal
  likelihood.
\newblock {\em YouTube}.

\bibitem[\protect\citename{Lepage, }1978]{lepage1978new}
Lepage, GP. 1978.
\newblock A new algorithm for adaptive multidimensional integration.
\newblock {\em Journal of Computational Physics}, {\bf 27}(2), 192--203.

\bibitem[\protect\citename{Neal, }2001]{neal2001annealed}
Neal, RM. 2001.
\newblock Annealed importance sampling.
\newblock {\em Statistics and Computing}, {\bf 11}(2), 125--139.

\bibitem[\protect\citename{O'Hagan, }1987]{o1987monte}
O'Hagan, A. 1987.
\newblock Monte Carlo is fundamentally unsound.
\newblock {\em The Statistician},  247--249.

\bibitem[\protect\citename{Rasmussen \& Ghahramani,
  }2003]{rasmussen2003bayesian}
Rasmussen, CE, \& Ghahramani, Z. 2003.
\newblock Bayesian Monte Carlo.
\newblock {\em Advances in Neural Information Processing Systems},  505--512.

\bibitem[\protect\citename{Skilling, }2006]{skilling2006nested}
Skilling, J. 2006.
\newblock Nested sampling for general Bayesian computation.
\newblock {\em Bayesian Analysis}, {\bf 1}(4), 833--859.

\end{thebibliography}

\end{document}